%% file: main.tex
\documentclass[runningheads]{llncs}
\pdfoutput=1
\usepackage{graphicx}
\usepackage{comment}
\usepackage{amsmath,amssymb} %
\usepackage{color}
\usepackage{adjustbox}
\usepackage[caption=false]{subfig}
\usepackage{makecell}
\usepackage{multicol}
\usepackage{multirow}
\usepackage{textcomp}
\usepackage{hyperref}
\usepackage[flushleft]{threeparttable}
\usepackage{url}
\usepackage[final]{pdfpages}
\usepackage{pax}

\begin{document}
\pagestyle{headings}
\mainmatter
\def\ECCVSubNumber{691}  %

  \title{Holopix50k: A Large-Scale In-the-wild Stereo Image Dataset} %

\titlerunning{Holopix50k}

\author{Yiwen Hua\orcidID{0000-0003-2334-7157} \and
Puneet Kohli\orcidID{0000-0002-4817-8436} \and
Pritish Uplavikar\thanks{indicates equal contribution}\orcidID{0000-0001-7187-8652} \and
Anand Ravi\textsuperscript{*}\orcidID{0000-0002-3656-601X} \and
Saravana Gunaseelan\orcidID{0000-0002-3315-7464} \and
Jason Orozco\orcidID{0000-0002-0638-3407} \and
Edward Li\orcidID{0000-0003-2651-1100}}
\authorrunning{Y. Hua et al.}
\institute{{Leia Inc.}}
\maketitle

\vspace{-0.25em}
\begin{figure}
    \centering
    \includegraphics[width=0.825\textwidth]{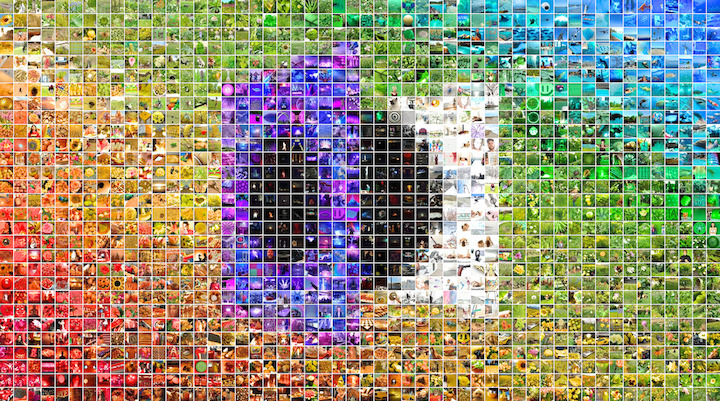}
    \caption{Samples from the Holopix50k stereo image dataset}
    \label{fig:mosaic}
\end{figure}
\vspace{-1.25em}

\input{sections/0_abstract}
\input{sections/1_introduction}
\input{sections/2_related_work}
\input{sections/3_data_acquisition_preprocessing}
\input{sections/4_cross_dataset_comparison}

\input{sections/5_experimentation/5_experimentation}
\input{sections/6_future_work}
\input{sections/7_conclusion}

\clearpage
\bibliographystyle{splncs04}
\bibliography{reference.bib}



\section*{Supplementary material}

# Supplementary Material

## Holopix50k: A Large-Scale In-the-wild Stereo Image Dataset

Yiwen Hua, Puneet Kohli, Pritish Uplavikar*, Anand Ravi*, Saravana Gunaseelan, Jason Orozco, and Edward Li

Leia Inc.

**Abstract.** This supplementary material adds discussions and results that were not included in the main paper. We additionally show results from experiments that were omitted from the main paper to ensure brevity.

## 1 About Holopix

Holopix was created in 2018 by Leia Inc. as a Lightfield image-sharing social media platform. It was launched along with the Hydrogen One Lightfield mobile device, created in partnership with the industrial camera manufacturer RED and pre-loaded on these devices. The user experience of this app is similar to other mobile photo-sharing platforms, such as Instagram, where users upload photos to the platform for others to view and interact within feeds. The app curates the top images to showcase to other users on the platform. The key difference compared to traditional image-sharing platforms is that on Holopix, the images uploaded are multi-view images or are converted to multi-view images by the Holopix platform. The multi-view images allow for both stereoscopic depth and a parallax effect depending on the user's device and viewing method. The conversion to multi-view is usually done by disparity estimation, followed by novel view-synthesis, and finally, image in-painting.

## 2 Data Filtering

### 2.1 Vertical Disparity

As our dataset is collected with cameras that are horizontally aligned, the stereo pairs are expected only to have horizontal disparity. Hence images with even slight vertical disparity can cause severe failures in matching algorithms such as stereo block-matching. Figure 1 shows the results of applying a horizontal stereo block-matching algorithm to an image pair with and without vertical disparity. As it is visible, introducing vertical disparity causes complete failure of the algorithm.

---

* indicates equal contribution

The reason for having vertical disparity could range from a variety of factors such as physically misaligned cameras, calibration errors, or synchronization issues. We filter out stereo pairs with any vertical disparity by calculating the vertical phase correlation between the left and right image pair.

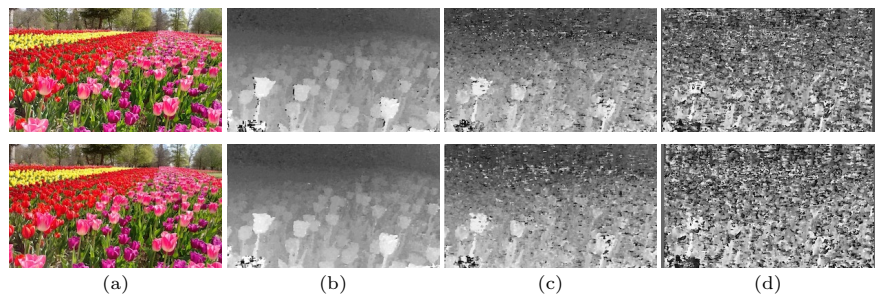

**Fig. 1.** (a) Input stereo pair (b) Disparity maps calculated using stereo block-matching (c) Disparity maps calculated with 2 pixels of vertical offset (d) Disparity maps calculated with 5 pixels of vertical offset. Top row corresponds to the left image and bottom row corresponds to the right image.

## 2.2   Disparity-based Filtering

We filtered the dataset based on the disparity maps generated from our stereo neural network. In cases such as extremely far-away scenes, where there is no stereo difference between left and right images, we would see apparent artifacts in the generated disparity map. There are other cases where an object is extremely close to the stereo cameras, causing significant visual differences in the stereo pair. A typical example of this is a photographer's finger partially covering the cameras. Additionally, as users on Holopix are free to upload stereo pairs from any source, there are also a variety of computer-generated pairs that do not exhibit stereo characteristics, or completely invalid stereo pairs where the left and right image do not match. These pairs are also flagged in our disparity-based filtering process and subsequently removed from the dataset. Note that the process involves human inspection of the stereo pairs and the disparity images to determine what should be filtered out. We also applied a low-standard deviation filter to filter out flat scenes that are generally texture-less or monochromatic. Figure 2 shows some examples of filtered stereo pairs and the calculated disparity map that was used to determine that these should be filtered out.

## 2.3   Dataset Indexes

We release 49,368 image pairs, which are indexed from posts on Holopix. If a user decides to delete their post from Holopix, the previously indexed image will

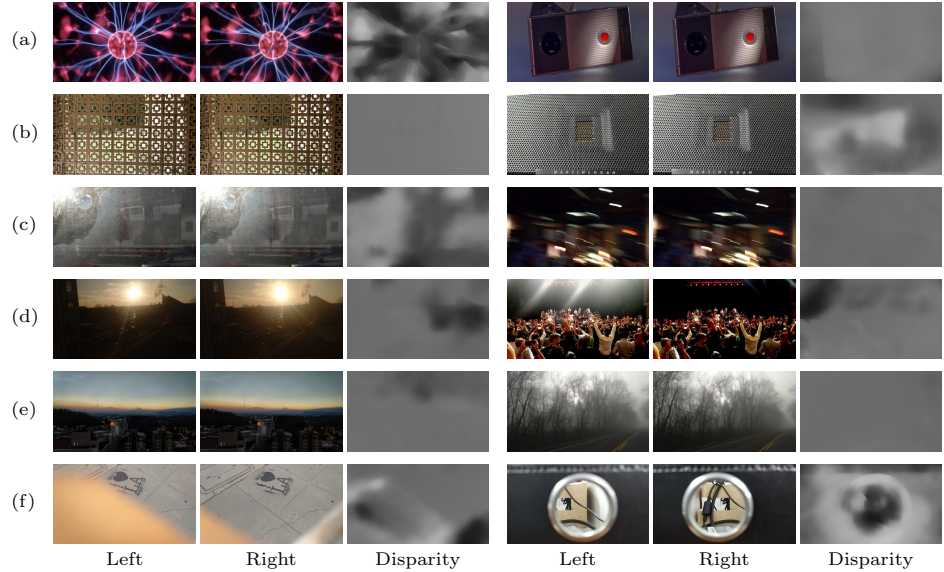

Left            Right           Disparity                    Left            Right           Disparity

**Fig. 2.** Examples of images filtered out from the Holopix50k dataset. The original stereo pair and the left disparity image generated by our stereo disparity model is shown. (a) Computer-generated images, (b) Images that are flat or have repeating textures, (c) Images with strong lens flare or motion blur, (d) images with significant lighting differences, (e) images of far-away scenes, (f) images of extremely close objects

no longer be available as a part of our dataset. As a result, the available portion of the dataset may shrink over time. This is similar to other datasets that are indexed from user-generated sources such as Flickr or Youtube.

## 3   Dataset Comparison

In addition to the image metric comparisons in the main paper, we provide additional dataset property comparisons in Table 1. Apart from being the largest stereo image dataset, we see that our dataset contains both indoor and outdoor in-the-wild scenes similar to Flickr1024 and WSVD. In contrast, the KITTI dataset contains only outdoor street scenes, and the Middlebury dataset contains indoor scenes in a laboratory setting. ETH3D contains varied scenes that consist of a limited number of indoor and outdoor imagery of buildings, parking lots, and some natural scenes. Our dataset also contains a mixture of both portrait and landscape orientation images unlike other datasets, except Flickr1024. Similar to the other in-the-wild stereo datasets, we do not provide ground truth disparity.

Table 2 shows the metric scores for the randomly sampled train, test, and validation subsets of our datasets. It can be observed that all the subsets show sim-

**Table 1.** Comparison of properties of popular stereo image datasets

| Dataset | Image Pairs | Resolution | Ground Truth | Orientation | Setting |
|---------|-------------|------------|--------------|-------------|---------|
| *KITTI 2012* | 389 | 0.46 Mpx | ✓ | Landscape | Street scenes |
| *KITTI 2015* | 400 | 0.47 Mpx | ✓ | Landscape | Street scenes |
| *Middlebury* | 65 | 3.59 Mpx | ✓ | Landscape | Laboratory |
| *ETH 3D* | 47 | 0.38 Mpx | ✓ | Landscape | Varied |
| *Flickr1024* | 1024 | 0.73 Mpx | | Landscape + Portrait | In-the-wild |
| *WSVD* | 10250 | 0.77 Mpx | | Landscape | In-the-wild |
| *Holopix50k* | 49368 | 0.74 Mpx | | Landscape + Portrait | In-the-wild |

**Table 2.** Mean and standard deviation of main characteristics of Holopix50k train, test and validation subsets

| Dataset | Image Pairs | Resolution (↑) | Entropy (↑) | BRISQUE (↓) | SR-metric (↑) | ENIQA (↓) |
|---------|-------------|----------------|-------------|-------------|---------------|-----------|
| *Holopix50k* | 49368 | 0.74 (±0.30) Mpx | 7.38 (±0.48) | 19.49 (±11.81) | 7.28[1](±1.24) | 0.174 (±0.125) |
| *Holopix50k (Train)* | 41954 | 0.74 (±0.30) Mpx | 7.38 (±0.48) | 19.49 (±11.83) | 7.27[1](±1.24) | 0.175 (±0.125) |
| *Holopix50k (Test)* | 2471 | 0.73 (±0.31) Mpx | 7.38 (±0.50) | 19.52 (±11.94) | 7.30[1](±1.25) | 0.170 (±0.123) |
| *Holopix50k (Val)* | 4943 | 0.74 (±0.31) Mpx | 7.37 (±0.48) | 19.46 (±11.64) | 7.29[1](±1.23) | 0.175 (±0.124) |

For all metrics, higher values are better, except BRISQUE and ENIQA.
[1] Only *left* images from Holopix HD were used to compute SR-metric due to computation constraints. We assume trivial perceptual differences between *left* and *right* images.

ilar performance across all the metrics. Additionally, these scores are all nearly identical to the dataset as a whole, indicating a negligible selection bias.

## 4    Stereo Super-Resolution

In addition to training PASSRNet models from scratch with different amounts of samples from our Holopix50k dataset, we also fine-tuned the model pre-trained on the Flickr1024 dataset. The PSNR and SSIM values of these experiments are reported in Table 3. In contrast to the from-scratch results reported in the main paper, we see that with the pre-trained weights, even a small sample (1k) of our dataset boosts performance on a variety of test sets. When using the entire dataset, the performance of the fine-tuned and the from-scratch models are quite similar.

Figure 3 graphically shows the change in PSNR and SSIM scores with respect to the number of training images, across our different PASSRNet models, as well as the original model trained on Flickr1024. We note that fine-tuning generally

**Table 3.** PSNR (dB) and SSIM values from training PASSRNet on various datasets for 4x SR with 80 training epochs

| Dataset | KITTI2015 (Test) | Middlebury (Test) | Flickr1024 (Test) | ETH3D (Test) | Holopix50k (Test) |
|---|---|---|---|---|---|
| *Flickr1024 (Train)* | 25.62 / 0.79 | 36.55 / 0.94 | **23.25 / 0.72** | 31.94 / 0.88 | 26.96 / 0.79 |
| *Holopix50k (1K) FT* | 25.46 / 0.78 | 36.25 / 0.94 | 22.93 / 0.70 | 32.10 / 0.88 | 27.14 / 0.80 |
| *Holopix50k (5K) FT* | 25.66 / 0.79 | <u>36.58 / 0.94</u> | 23.05 / 0.71 | 32.29 / 0.89 | 27.38 / 0.81 |
| *Holopix50k (10K) FT* | <u>25.68 / 0.79</u> | 36.65 / 0.94 | 23.09 / 0.71 | 32.37 / 0.89 | 27.42 / 0.81 |
| *Holopix50k (Train) FT* | 25.60 / 0.79 | **36.86 / 0.94** | <u>23.15 / 0.71</u> | 32.45 / 0.89 | **27.58 / 0.81** |
| *Holopix50k (Train)* | **25.80 / 0.80** | **36.86 / 0.94** | 23.10 / 0.71 | **32.56 / 0.89** | 27.51 / 0.81 |

Note that *FT* above stands for Fine-tuned. **Bold** values indicate best scores, <u>underlined</u> values indicate second best scores.

produces better results for a smaller number of samples. As we increase the training data, the fine-tune and from-scratch models converge. As expected, using more training data improves results across the board.

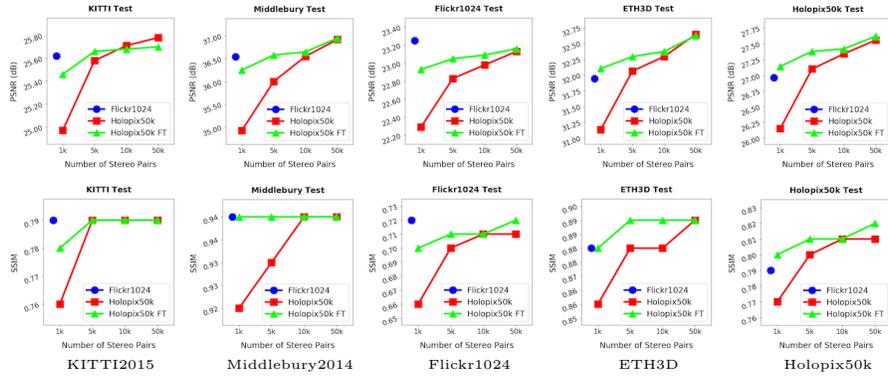

**Fig. 3.** PSNR and SSIM values for 4x SR tasks obtained by PASSRNet models trained on different number of stereo pairs

In Figure 4, we show additional qualitative results of the PASSRNet model trained on Holopix50k, compared to other super-resolution methods. In-line with the results we show in the main paper, we observe sharper details and higher quality textures in samples super-resolved with our model.

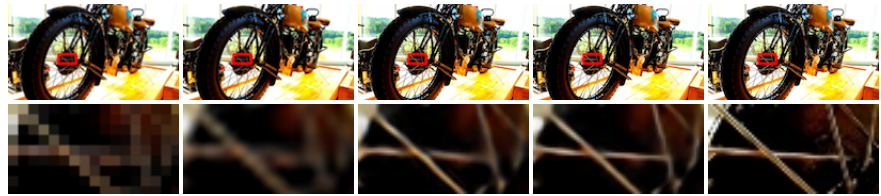

| Input<br>Holopix50K | Bicubic<br>19.59/0.69 | Flickr1024[*]<br>21.74/0.79 | Holopix50k[*]<br>22.32/0.82 | Ground Truth |

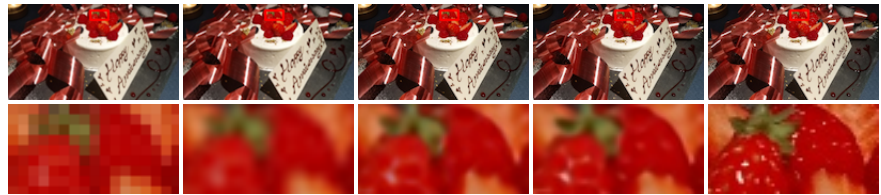

| Input<br>Holopix50K | Bicubic<br>24.92/0.78 | Flickr1024[*]<br>27.45/0.85 | Holopix50k[*]<br>28.44/0.87 | Ground Truth |

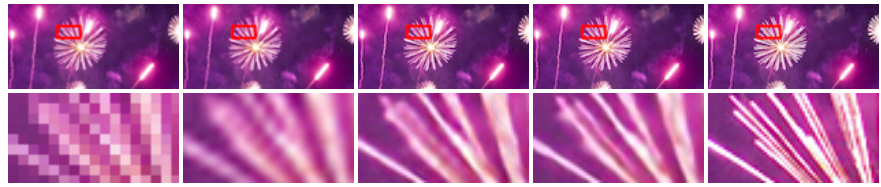

| Input<br>Flickr1024 | Bicubic<br>26.55/0.86 | Flickr1024[*]<br>27.47/0.87 | Holopix50k[*]<br>27.79/0.88 | Ground Truth |

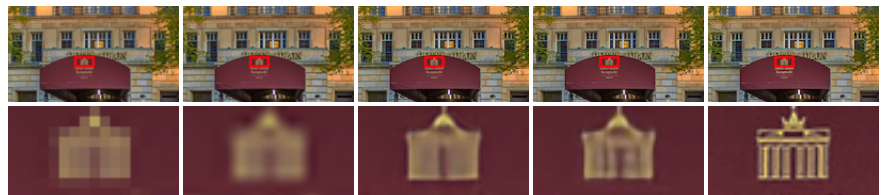

| Input<br>Flickr1024 | Bicubic<br>21.35/0.58 | Flickr1024[*]<br>23.06/0.70 | Holopix50k[*]<br>22.80/0.70 | Ground Truth |

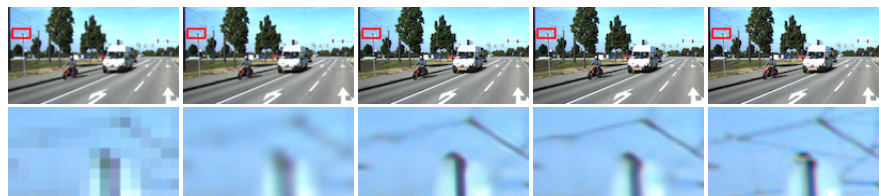

| Input<br>KITTI2015 | Bicubic<br>25.72/0.82 | Flickr1024[*]<br>27.30/0.82 | Holopix50k[*]<br>27.38/0.82 | Ground Truth |

[*] Refers to PASSRNet trained on the respective datasets

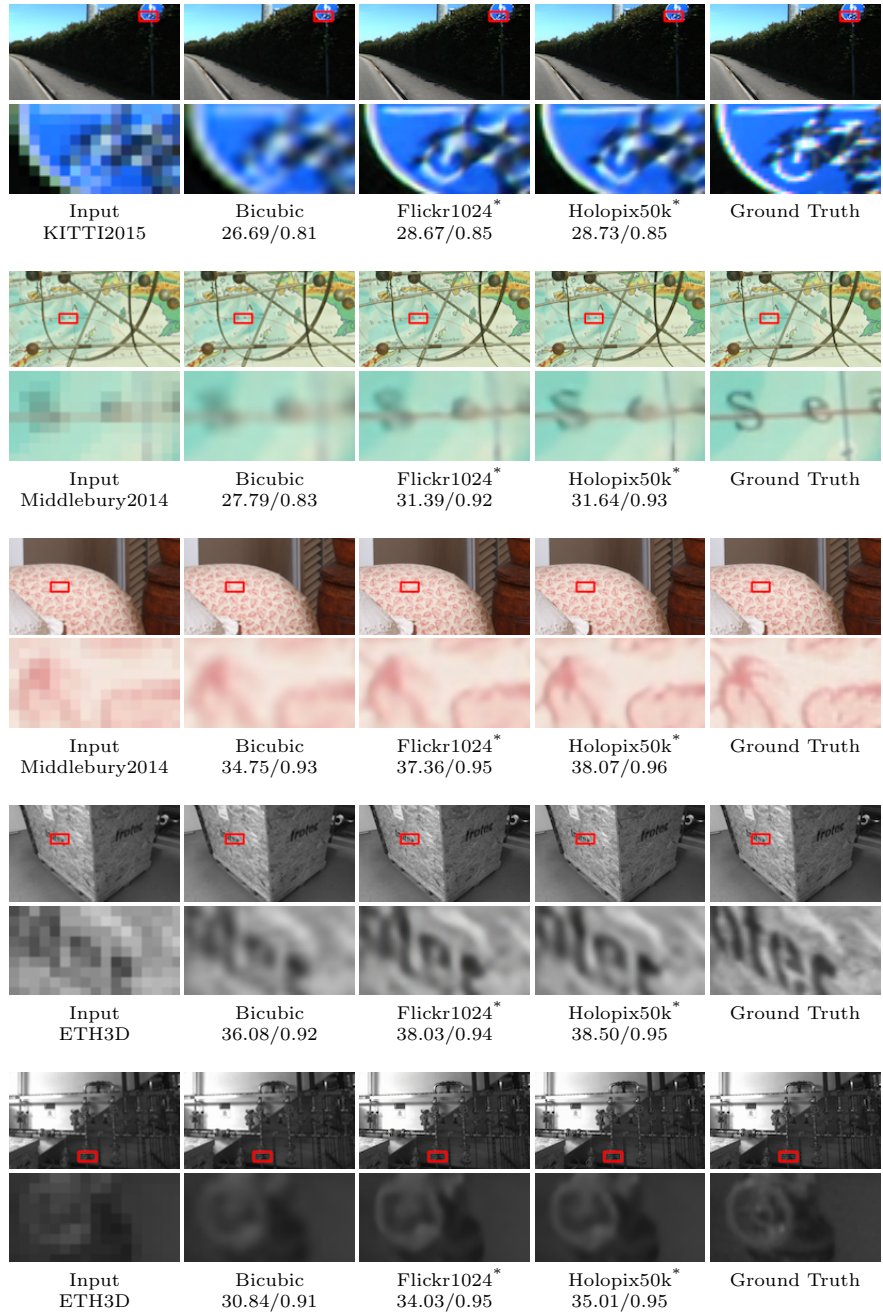

**Fig. 4.** Additional results of SR methods for 4x SR on samples from a variety of datasets.

## 5   Self-Supervised Depth Estimation

### 5.1   Implementation Details

In the original implementation of Monodepth2, the authors use a reprojection-based image warping technique, which requires camera pose information. As the stereo pairs in our dataset are already rectified, we modify the image warping to use a disparity-based backward warping technique instead, which does not require pose information. We additionally skip the disparity to depth conversion during training as we do not have ground truth depth to bound our depth scale. Finally, we change the smoothness term $\lambda$ to 1.0 as we empirically found that it generates the best results for our experiments.

### 5.2   Additional Results

We report the fine-tune results of training the Monodepth2 model in our main paper, as we focus on the generalization our dataset empowers. Specifically, in the self-supervised case, the generalization of a model trained on street scenes (KITTI) to laboratory (Middlebury), and synthetic (MPI Sintel) scenes.

**Table 4.** Disparity comparison of running Monodepth2 models on the Middlebury visual benchmark and MPI Sintel test tests, respectively

| Dataset | Middlebury Test | | | | MPI Sintel Test | | | |
|---|---|---|---|---|---|---|---|---|
| | Abs Rel | Sq Rel | RMSE | RMSE log | Abs Rel | Sq Rel | RMSE | RMSE log |
| *KITTI* | 0.590 | 15.968 | 20.891 | 0.308 | 0.337 | 31.726 | 41.597 | 0.173 |
| *Holopix50k (1K)* | 0.420 | 7.502 | 14.268 | 0.23 | 0.236 | 17.040 | **25.279** | 0.113 |
| *Holopix50k (5K)* | 0.422 | 7.585 | 14.302 | 0.204 | 0.240 | 18.075 | 25.747 | 0.114 |
| *Holopix50k (10K)* | 0.420 | 7.501 | 14.224 | 0.202 | 0.236 | <u>17.018</u> | 25.335 | 0.113 |
| *Holopix50k (Train)* | **0.414** | **7.269** | **14.032** | **0.200** | <u>0.235</u> | 17.155 | 25.357 | **0.113** |
| *Holopix50k (Train) FT* | <u>0.417</u> | <u>7.373</u> | <u>14.112</u> | <u>0.201</u> | **0.235** | **16.810** | <u>25.302</u> | <u>0.113</u> |

**Bold** values indicate best scores, <u>underlined</u> values indicate second-best scores. For all metrics, lower values are better. *FT* refers to fine-tuned model originally trained on the KITTI dataset.

In addition to fine-tuning the Monodepth2 model that was originally trained on the KITTI dataset, we also train the model from scratch with different amounts of samples from our dataset. The results are reported in Table 4. Interestingly, we observe that the model does not always perform better with the increase in the amount of training data. For example, the 1k model has better results than the 5k model. We believe this is due to the challenging images our dataset contains for the monocular depth estimation task, and more data is required for the model to learn the distribution of our data. We also observe that

the from-scratch models out-perform the fine-tune models. We theorize this is because of the difference in distribution between KITTI and Holopix50k datasets, and perhaps the pre-trained model serves as a less-than-ideal initialization for the Holopix50k data.

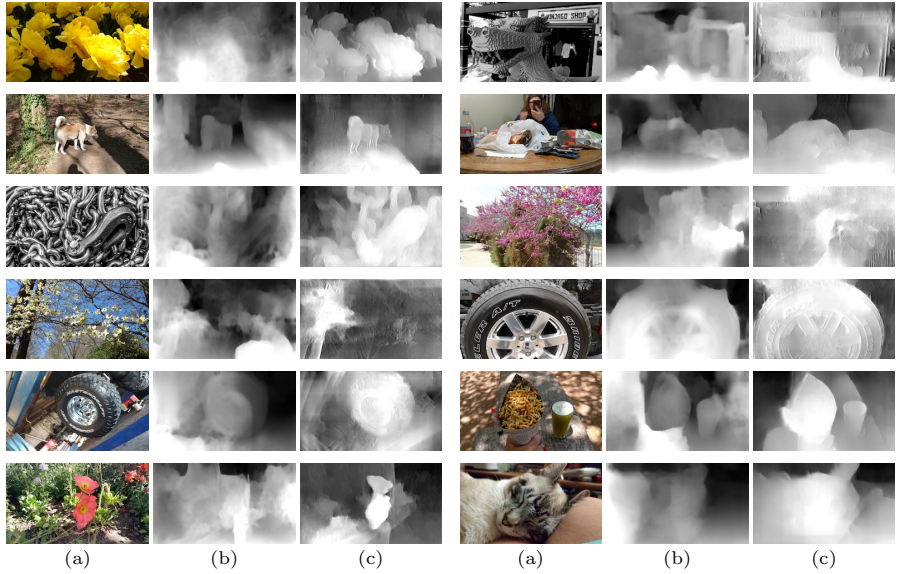

**Fig. 5.** Disparity maps produced by Monodepth2 models on samples from the Holopix50k test set. (a) Input image, (b) Trained on KITTI, (c) Trained on Holopix50k

Figure 5 shows additional qualitative results of the Monodepth2 models on samples from the Holopix50k test set. Similar to the results from Middlebury and MPI Sintel test sets, the results are sharper and respect edges better when trained on Holopix50k. We do not show ground truth disparities as we do not have the ground truth disparities for our dataset.

# 6    More Disparity Results

In Figure 6 we show additional results from running our various disparity estimation models. Note that strong light reflections are modeled as Lambertian effects. Hence disparity maps should display far back (dark) values for light reflections.

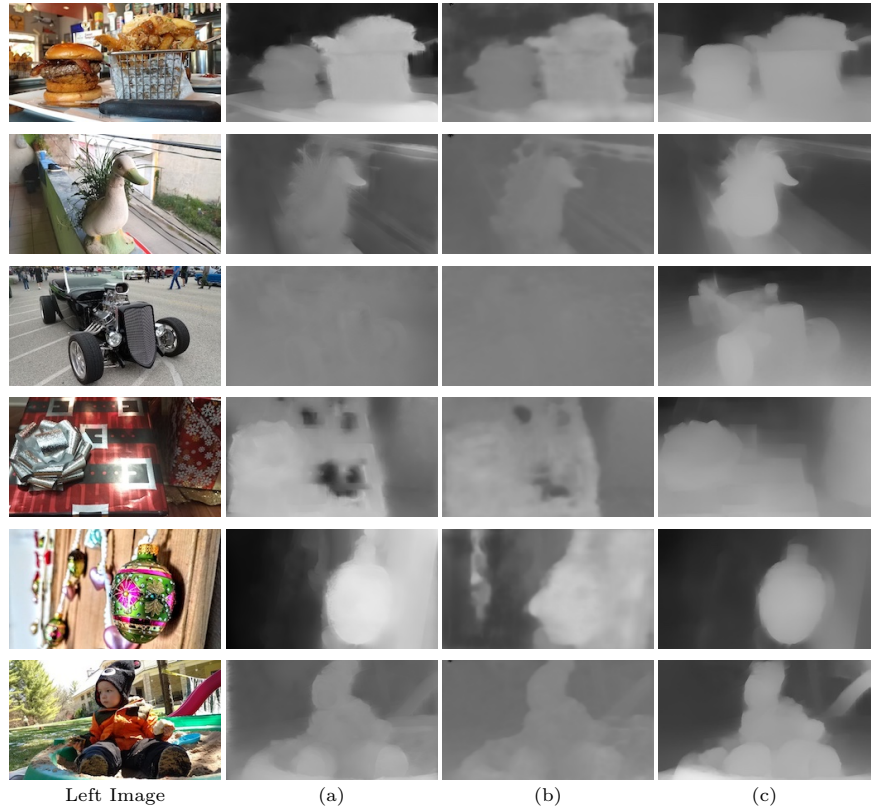

Left Image                (a)                (b)                (c)

**Fig. 6.** Left image disparity maps produced by our models on samples from Holopix50k. (a) shows left disparity from our stereo disparity model, (b) shows result of our real-time disparity estimation, and (c) shows relative depth from our monocular depth estimation model

# 7   Dataset Examples

In Figure 7 we show sample stereo pairs present in the Holopix50k dataset.

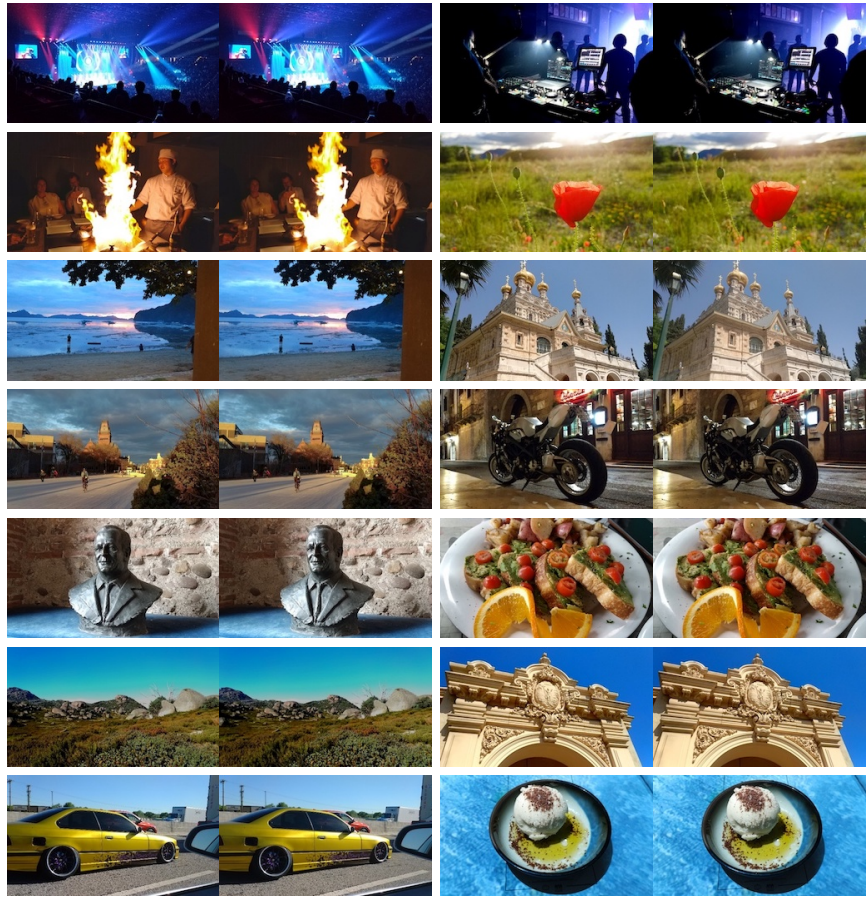

**Fig. 7.** Stereo images present in the Holopix50k dataset


\end{document}

%% file: sections/0_abstract.tex
\begin{abstract}
With the mass-market adoption of dual-camera mobile \linebreak phones, leveraging stereo information in computer vision has become increasingly important. Current state-of-the-art methods utilize learning-based algorithms, where the amount and quality of training samples heavily influence results. Existing stereo image datasets are limited either in size or subject variety. Hence, algorithms trained on such datasets do not generalize well to scenarios encountered in mobile photography. We present Holopix50k, a novel in-the-wild stereo image dataset, comprising 49,368 image pairs contributed by users of the Holopix\texttrademark\ mobile social platform. In this work, we describe our data collection process and statistically compare our dataset to other popular stereo datasets. We experimentally show that using our dataset significantly improves results for tasks such as stereo super-resolution and self-supervised monocular depth estimation. Finally, we showcase practical applications of our dataset to motivate novel works and use cases. The dataset is available at \url{http://github.com/leiainc/holopix50k}.

\keywords{Stereo vision \and stereo image dataset \and 3D photography \and stereo super-resolution \and disparity estimation}
\end{abstract}

%% file: sections/1_introduction.tex
\section{Introduction}

Most mobile phones now come with two or more cameras, enabling consumer applications such as artificial depth of field (portrait mode), AI-based photo optimization, and 3D photography \cite{facebook_3D}. Through dual-cameras, mobile phones can perceive depth and lighting of a scene using complementary information from the stereo image pairs. Various applications utilize stereo imagery such as Holopix\footnote{Holopix\texttrademark. The leading mobile social network for sharing lightfield imagery (\url{https://www.holopix.com}).} to produce immersive 3D Photography experiences. More recently, larger platforms such as Facebook and Snap Inc. have also demonstrated unique and exciting user experiences enabled via stereo imagery.

Advancements in self-driving cars and robotics technologies with \linebreak multi-camera configurations have accelerated research and development of computer vision algorithms \cite{monodepth2}\cite{passrnet2019longguang}, many of which utilize additional stereo information \cite{stereo_vision_1}\cite{stereo_vision_2}\cite{stereo_vision_3}. The mass-market adoption of dual-cameras in mobile phones has led to an increased need for these algorithms to work in casual in-the-wild scenarios encountered during mobile photography. A majority of state-of-the-art methods in this domain are based on deep learning methods, where accuracy and quality scale with the amount of data available for training. Current datasets either cover a limited subset of real-world scenarios \cite{kitti2012}\cite{cordts2016cityscapes} or are taken in a laboratory setting \cite{middlebury}\cite{dagobert2018production}. Further, there is an increased need for a large-scale stereo dataset representative of the diversity of real-world scenarios to enable generalization. Though recent works have been able to collect stereo pairs \cite{wsvd}\cite{flickr1024} from various internet platforms, they may not be diverse or extensive enough to represent in-the-wild scenarios captured from mobile devices. 

In this work, we present a new large-scale stereo image dataset that is completely user-generated and captured mostly from mobile phone cameras in-the-wild. In our novel \textit{Holopix50k} dataset, we present 49,368 (roughly 50k) rectified stereo image pairs collected from the mobile social media platform Holopix. This is larger than comparable stereo image datasets by an order of magnitude. To our knowledge, this is the first large-scale stereo image dataset collected from a mobile-based social platform where users curate and upload stereo images. Figure \ref{fig:mosaic} shows samples from the dataset demonstrating the diversity of scenes present, arranged to depict the Holopix platform's logo. 

In the next section, we explore existing stereo datasets and their applications to stereo vision tasks. In the following sections, we explore our data collection and filtering process. Further, we compare our dataset with existing stereo datasets on various metrics. Next, we demonstrate how our dataset helps improve existing state-of-the-art methods for stereo super-resolution and self-supervised monocular depth estimation tasks. Finally, we showcase the qualitative results of various disparity models we have trained using our dataset to motivate further novel works and use cases.
\\\\
Summarizing our key contributions below,
\begin{enumerate}
    \item Our new dataset, Holopix50k, is the largest in-the-wild stereo image dataset to date with 49,368 high-quality rectified stereo image pairs covering a wide variety of realistic scenarios found in mobile photography.
    \item Our experimental results demonstrate that our dataset improves a variety of stereo vision tasks in both quantitative and qualitative performance. 
\end{enumerate}

%% file: sections/2_related_work.tex
\section{Related Work}

\subsection{Stereo Datasets}

\subsubsection{Real-world Datasets.}
There are a variety of real-world stereo datasets currently available. Some of the more popular ones, such as KITTI \cite{kitti2012}\cite{kitti2015}, Middlebury \cite{middlebury}, and the NYU Indoor \cite{Silberman:ECCV12} datasets, all provide ground truth depth or disparities for the stereo scenes. However, these datasets either have a limited number of images, or the scenes are for a specific domain. For example, the KITTI datasets were mainly developed for self-driving vehicle use-cases, while the few scenes in Middlebury are all captured in a laboratory setting. Other real-world stereo datasets include Make3D \cite{saxena2008make3d}, ETH3D \cite{eth3d}, CMLA \cite{dagobert2018production}, and Cityscapes \cite{cordts2016cityscapes} --- each focused on a specific domain. More recent datasets such as Flickr1024 \cite{flickr1024} and WSVD \cite{wsvd} provide more diverse scenes, however, Flickr1024 is relatively small compared to our dataset and WSVD images score significantly lower on quality metrics, as shown in Table \ref{table:cross_dataset_metrics}. While our dataset also comprises real-world scenarios, we demonstrate the superiority of our dataset in terms of the number and diversity of images present, as well as various metric scores.

\subsubsection{Synthetic Datasets.}

The challenge with real-world stereo datasets is collecting ground truth information at-scale that covers a large variety of complex scene types, lighting effects, and motion scenarios. It is also difficult to collect data with different camera baselines and depth ranges. Recent works introduce synthetic datasets that take advantage of the rise of efficient, high-quality rendering techniques. Examples include MPI Sintel \cite{butler2012mpi}, SceneFlow \cite{MIFDB16}, UnrealStereo \cite{zhang2018unrealstereo}, and the 3D Ken Burns \cite{niklaus20193d} datasets. The challenge with such datasets is that techniques based on these datasets still suffer from the domain shift problem \cite{DBLP:journals/corr/abs-1711-06969} when adapting to real-world scenarios. Furthermore, significant effort may be required to cover the diversity of real-world scenarios using synthetic techniques. We envision our dataset improving the robustness and generalization capabilities of learning-based models trained on these synthetic datasets by providing a large distribution of real-world scenarios. 

\subsection{Stereo Image Tasks}

\subsubsection{Stereo Disparity Estimation.} 

Traditional stereo disparity estimation algorithms rely on block-matching methods \cite{sgbm}. More recently, there has been a significant interest in learning-based techniques due to advances in neural networks \cite{Zhang2019GANet}\cite{Yin_2019_CVPR}\cite{dispnet}\cite{khamis2018stereonet}. Most learning-based techniques require ground-truth disparity (or depth) for supervision. Since most datasets with ground-truth disparity are limited to specific domains, recent works use self-adaption approaches to estimate disparity \cite{Tonioni_2017_ICCV}\cite{Tonioni_2019_CVPR}\cite{pang2018zoom}. The diversity of our dataset can aid in the generalization of these unsupervised techniques. 

\subsubsection{Monocular Depth Estimation.} Monocular depth estimation aims to estimate the depth of a scene from a single image \cite{mono_disparity_1}\cite{mono_disparity_2}. This is an ill-posed problem as image textures do not directly correspond to depth \cite{DBLP:journals/corr/abs-1901-09402_mono_depth}. Monocular techniques exist using self-supervision with a stereo input, and some are entirely unsupervised \cite{mono_disparity_1}\cite{unsupervised_mono_disparity_2}. Similar to stereo disparity tasks, the datasets used are limited in either domain or size. Hence, these techniques fail to generalize to tasks found in-the-wild. Our dataset has a large variety of diverse scenarios at a scale no other dataset currently provides, which significantly improves the generalizability of models trained for these tasks.

\subsubsection{Stereo Super-Resolution.} Super-resolution is a well-known task in computer vision that aims to construct a high-resolution image from their low-resolution versions \cite{SingleSuperResolution2}\cite{SingleSuperResolution1}\cite{SingleSuperResolution3}. Stereo super-resolution techniques aim to incorporate a second low-resolution image to increase the quality of the super-resolved high-resolution image. Recently, there has been an increased interest in learning-based stereo super-resolution methods \cite{passrnet2019longguang}\cite{stereosr}\cite{StereoSuperResolution1} due to the advancement in stereo photography on mobile devices. The sheer size of our dataset is well-suited to improve the results of stereo super-resolution.
\\
\\
There are a variety of other stereo vision tasks such as stereoscopic style transfer \cite{stereo_style_transfer_1}\cite{stereo_style_transfer_2}\cite{pk2020gpustyle}, flow estimation \cite{flow_estimation_1}\cite{flow_estimation_2}, and stereo segmentation \cite{stereo_segmentation}, which could benefit from a large-scale dataset such as ours. We do not demonstrate the improvements that our dataset brings to such tasks in this work.

%% file: sections/3_data_acquisition_preprocessing.tex
\section{The Holopix50k Dataset}

Holopix50k is the largest in-the-wild stereo image dataset crowd-sourced from a social media platform to date, containing 49,368 stereo pairs (98,736 images in total). In this section, we discuss our data collection method and also shed light on the diversity of content present in our dataset.

\subsection{Data Acquisition and Processing}

The Holopix platform (visualized in Figure \ref{fig:holopix}) is the only major social media platform built specifically for sharing 3D Photography where multiple viewpoints of a scene are captured and viewed with a parallax effect \cite{facebook_3D}. Combined with Lightfield displays, such as those made by Leia Inc. \cite{Fattal2013}, there is an added depth \cite{howard1995binocular} and multi-view parallax effect \cite{dogdson2006autostereoscopic} that enhances the experience. Due to Holopix's unique position as the first Lightfield-enabled social media application, it has attracted a significant user-base who regularly upload photos to the platform. The platform takes in two or more input views and renders a multi-view image on Lightfield devices, or a motion-based animation on regular devices. 

\input{tables/fig_hydrogen_holopix}

Statistically, the vast majority of images on Holopix are captured on the RED Hydrogen One\footnote[1]{RED Hydrogen One. \url{https://en.wikipedia.org/wiki/Red_Hydrogen_One}} mobile phone, which is one of the first consumer-grade Lightfield devices, offering a 4-view display. The stereo camera application on this device rectifies images and converges on the mean disparity of the scene. Note that in works such as Flickr1024 \cite{flickr1024}, the stereo pairs are cropped to re-converge the stereo pairs at infinity, as a pre-processing step. To maintain the initially captured pair as-is, we do not re-converge the images via cropping.

\subsubsection{Data Collection.}

We initially collect about 70k stereo image pairs from the Holopix platform that were suitable for use in the dataset. In the case of mismatching resolution in the left and right stereo images, we downscale the larger of the two images to match the resolution of the smaller image for consistency. The downscaling is done by first applying Gaussian smoothing to the image, followed by bicubic interpolation, as Holopix also used this method to downscale the images.

\subsubsection{Removing Vertical Disparity.}

As our dataset originates from cameras that are horizontally aligned, the stereo pairs are expected only to have horizontal disparity. As such, stereo pairs with even slight vertical disparity can cause severe failures in stereo algorithms such as stereo block-matching. \cite{phaseCorrelate2}\cite{phaseCorrelate1}. After filtering out stereo pairs with vertical disparity, we are left with roughly 60k images in our dataset.

\subsubsection{Disparity-based Filtering.}

Given that most images from our dataset are captured on the Hydrogen One, which has a small stereo baseline of 12mm for the rear camera and 5mm for the front camera, we expect certain scene conditions to produce limited stereo information. We filter out these images by analyzing the disparity maps produced by our stereo disparity network (see Section \ref{sec:internal_training}) and removing the images that produce extreme disparity failures. By doing this, we bias our dataset towards scenes which have useful disparity information. We provide more details in the supplemental material.

Our filtering process removes up to 95\% of image pairs with poor stereo characteristics, after which, our final dataset includes 49,368 stereo pairs.

\subsection{Dataset Exploration}

As an exploration of our dataset, we run the Mask-RCNN \cite{maskrcnn} object detector on the left image of each stereo pair. The detector we use is pre-trained on the MS COCO dataset \cite{coco}, containing 91 object categories for common objects. We retain only the bounding boxes with $\geq 0.7$ confidence. From the word-cloud representation in Figure \ref{fig:word_cloud}, we can see that there is a good mixture of common objects including people, animal species, plants, vehicles, furniture, electronics, and food types. As Holopix is a social media platform, it is expected that a majority of the images would contain humans, which indeed dominates our dataset with approximately 21k humans detected. 

\input{tables/fig_wordcloud}

\input{tables/fig_class_images}

Our dataset also contains both landscape as well as portrait orientation photography. Although a small portion of images are in portrait orientation (roughly 4\%), this adds additional diversity to the dataset in terms of unique content such as stereoscopic selfies. Figure \ref{fig:dataset_diversity} shows hand-picked samples from our dataset that demonstrate even more diversity in terms of scene categories.

%% file: tables/fig_hydrogen_holopix.tex
\begin{figure}[!htb]
    \centering
    \includegraphics[width=0.6\textwidth]{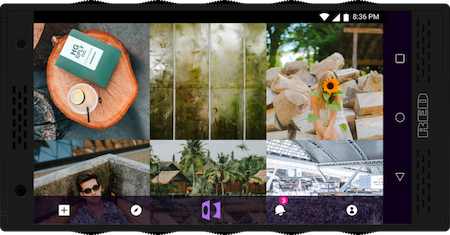}
    \caption{The Holopix Social Platform}
    \label{fig:holopix}
\end{figure}

%% file: tables/fig_wordcloud.tex
\begin{figure}[!htb]
    \centering
    \includegraphics[width=0.825\linewidth]{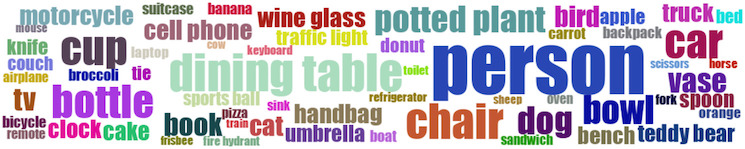}
    \caption{Word-cloud depicting common objects found in the Holopix50k dataset using the Mask-RCNN \protect{\cite{maskrcnn}} object detector trained on the COCO dataset \protect{\cite{coco}}}
    \label{fig:word_cloud}
\end{figure}

%% file: tables/fig_class_images.tex
\begin{figure}[!htb]
\centering
\begingroup

\setlength{\tabcolsep}{0.5pt}
    \scriptsize
    \begin{tabular}{ccccc@{\hskip0.05\linewidth}ccccc@{\hskip0.05\linewidth}ccccc}

    \subfloat{\includegraphics[width = 0.05\linewidth]{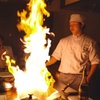}} &
    \subfloat{\includegraphics[width = 0.05\linewidth]{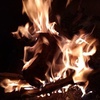}} &
    \subfloat{\includegraphics[width = 0.05\linewidth]{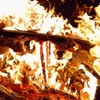}} &
    \subfloat{\includegraphics[width = 0.05\linewidth]{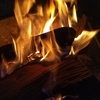}} &
    \subfloat{\includegraphics[width = 0.05\linewidth]{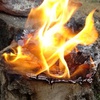}} &
    \subfloat{\includegraphics[width = 0.05\linewidth]{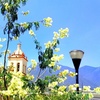}} &
    \subfloat{\includegraphics[width = 0.05\linewidth]{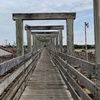}} &
    \subfloat{\includegraphics[width = 0.05\linewidth]{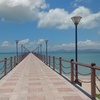}} &
    \subfloat{\includegraphics[width = 0.05\linewidth]{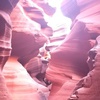}} &
    \subfloat{\includegraphics[width = 0.05\linewidth]{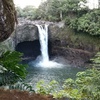}} &
    \subfloat{\includegraphics[width = 0.05\linewidth]{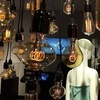}} &
    \subfloat{\includegraphics[width = 0.05\linewidth]{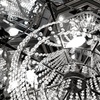}} &
    \subfloat{\includegraphics[width = 0.05\linewidth]{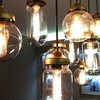}} &
    \subfloat{\includegraphics[width = 0.05\linewidth]{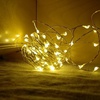}} &
    \subfloat{\includegraphics[width = 0.05\linewidth]{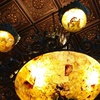}} \\
    \multicolumn{5}{c@{\hskip0.05\linewidth}}{fire} & \multicolumn{5}{c@{\hskip0.05\linewidth}}{landscape} & \multicolumn{5}{c}{lights} \\
    
    \subfloat{\includegraphics[width = 0.05\linewidth]{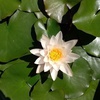}} &
    \subfloat{\includegraphics[width = 0.05\linewidth]{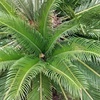}} &
    \subfloat{\includegraphics[width = 0.05\linewidth]{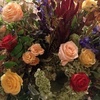}} &
    \subfloat{\includegraphics[width = 0.05\linewidth]{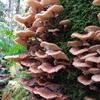}} &
    \subfloat{\includegraphics[width = 0.05\linewidth]{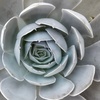}} &
    \subfloat{\includegraphics[width = 0.05\linewidth]{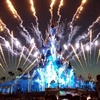}} &
    \subfloat{\includegraphics[width = 0.05\linewidth]{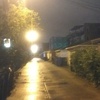}} &
    \subfloat{\includegraphics[width = 0.05\linewidth]{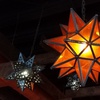}} &
    \subfloat{\includegraphics[width = 0.05\linewidth]{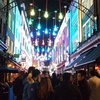}} &
    \subfloat{\includegraphics[width = 0.05\linewidth]{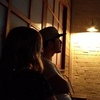}} &
    \subfloat{\includegraphics[width = 0.05\linewidth]{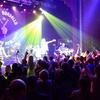}} &
    \subfloat{\includegraphics[width = 0.05\linewidth]{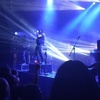}} &
    \subfloat{\includegraphics[width = 0.05\linewidth]{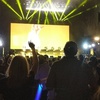}} &
    \subfloat{\includegraphics[width = 0.05\linewidth]{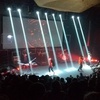}} &
    \subfloat{\includegraphics[width = 0.05\linewidth]{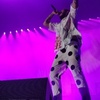}} \\
    \multicolumn{5}{c@{\hskip0.05\linewidth}}{nature} & \multicolumn{5 }{c@{\hskip0.05\linewidth}}{night} & \multicolumn{5}{c}{party} \\
    
    \subfloat{\includegraphics[width = 0.05\linewidth]{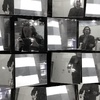}} &
    \subfloat{\includegraphics[width = 0.05\linewidth]{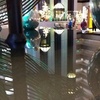}} &
    \subfloat{\includegraphics[width = 0.05\linewidth]{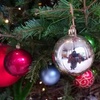}} &
    \subfloat{\includegraphics[width = 0.05\linewidth]{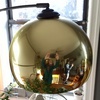}} &
    \subfloat{\includegraphics[width = 0.05\linewidth]{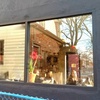}} &
    \subfloat{\includegraphics[width = 0.05\linewidth]{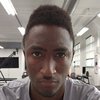}} &
    \subfloat{\includegraphics[width = 0.05\linewidth]{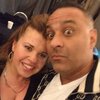}} &
    \subfloat{\includegraphics[width = 0.05\linewidth]{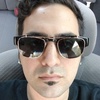}} &
    \subfloat{\includegraphics[width = 0.05\linewidth]{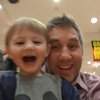}} &
    \subfloat{\includegraphics[width = 0.05\linewidth]{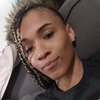}} &
    \subfloat{\includegraphics[width = 0.05\linewidth]{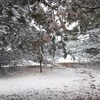}} &
    \subfloat{\includegraphics[width = 0.05\linewidth]{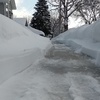}} &
    \subfloat{\includegraphics[width = 0.05\linewidth]{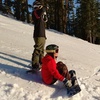}} &
    \subfloat{\includegraphics[width = 0.05\linewidth]{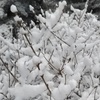}} &
    \subfloat{\includegraphics[width = 0.05\linewidth]{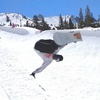}} \\
    \multicolumn{5}{c@{\hskip0.05\linewidth}}{reflections} & \multicolumn{5}{c@{\hskip0.05\linewidth}}{selfie} & \multicolumn{5}{c}{snow} \\
    
    \subfloat{\includegraphics[width = 0.05\linewidth]{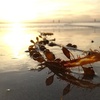}} &
    \subfloat{\includegraphics[width = 0.05\linewidth]{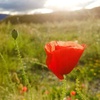}} &
    \subfloat{\includegraphics[width = 0.05\linewidth]{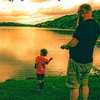}} &
    \subfloat{\includegraphics[width = 0.05\linewidth]{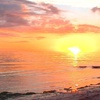}} &
    \subfloat{\includegraphics[width = 0.05\linewidth]{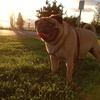}} &
    \subfloat{\includegraphics[width = 0.05\linewidth]{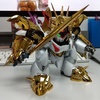}} &
    \subfloat{\includegraphics[width = 0.05\linewidth]{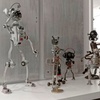}} &
    \subfloat{\includegraphics[width = 0.05\linewidth]{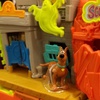}} &
    \subfloat{\includegraphics[width = 0.05\linewidth]{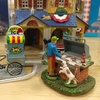}} &
    \subfloat{\includegraphics[width = 0.05\linewidth]{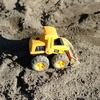}} &
    \subfloat{\includegraphics[width = 0.05\linewidth]{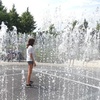}} &
    \subfloat{\includegraphics[width = 0.05\linewidth]{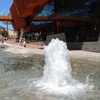}} &
    \subfloat{\includegraphics[width = 0.05\linewidth]{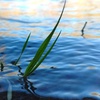}} &
    \subfloat{\includegraphics[width = 0.05\linewidth]{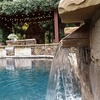}} &
    \subfloat{\includegraphics[width = 0.05\linewidth]{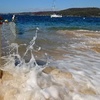}} \\
    \multicolumn{5}{c@{\hskip0.05\linewidth}}{sunset} & \multicolumn{5}{c@{\hskip0.05\linewidth}}{toys} & \multicolumn{5}{c}{water} \\
    \end{tabular}
    \endgroup
    \caption{Diversity of content found in the Holopix50k dataset}
    \label{fig:dataset_diversity}
\end{figure}

%% file: sections/4_cross_dataset_comparison.tex
\section{Comparison with Existing Datasets}

In this section, we show the results of statistical comparison of the Holopix50k dataset to other popular stereo datasets. Table \ref{table:cross_dataset_metrics} shows the results of comparison against KITTI2012 \cite{kitti2012}, KITTI2015 \cite{kitti2015}, Middlebury \cite{middlebury}, ETH3D \cite{eth3d}, Flickr1024 \cite{flickr1024}, and Web Stereo Video (WSVD) \cite{wsvd} datasets. Note that WSVD contains stereo videos collected from YouTube and Vimeo, filtered out into unique clips. Out of these, Flickr1024 and WSVD are ones that we consider to be in-the-wild datasets similar to ours, whereas the others are captured under specific environmental settings. We compare the amount of information present in the datasets using the entropy measure \cite{entropy}, and use BRISQUE \cite{brisque}, SR-metric \cite{srmetric}, and ENIQA \cite{eniqa}, in order to assess the no-reference perceptual quality of the datasets.

\input{tables/table_cross_dataset}

We divide our dataset into two subsets for analysis, {\it HD} and {\it SD}, containing the 720p and 360p images, respectively. From Table \ref{table:cross_dataset_metrics}, we observe that the lower resolution \textit{SD} subset performs well in the perceptual metrics, with the best score in SR-metric and second-best in BRISQUE, ENIQA, and entropy. Although the higher resolution \textit{HD} subset does not fare as well in these metrics, it has the second-highest resolution of all stereo datasets available, after Middlebury. Both of our subsets individually contain a larger number of images than any other dataset. 

When we analyze the dataset as a whole, Holopix50k is the largest stereo image dataset, beating the second-largest dataset (WSVD) by roughly five times, and the third-largest (Flickr1024) by almost fifty times. Compared to other datasets, our complete dataset has the highest SR-metric, which indicates that our dataset contains higher quality images with respect to human visual perception \cite{srmetric}. Our dataset also scores second-highest in entropy, which shows that, on average, the amount of information present in images of our dataset is high. 

These results showcase the superiority of our dataset on various fronts. First, our dataset has the most extensive selection of \textit{HD} stereo image pairs. Second, it has the most extensive selection of \textit{SD} stereo image pairs having top-2 scores across all metrics. Finally, as a whole, our dataset is orders of magnitudes larger than the rest, yet it performs relatively well in most perceptual metrics compared to the smaller datasets.

We additionally randomly split the dataset into train, test, and validation subsets based on an 85:5:10 ratio and for each subset to observe near-identical metric scores to the entire dataset, showing negligible selection bias (see supplemental material for details).

%% file: tables/table_cross_dataset.tex
\begin{table}
\begin{center}
\begin{adjustbox}{width=1\textwidth}
{\renewcommand{\arraystretch}{1.4}
\begin{threeparttable}
\caption{Mean and standard deviation of main characteristics of popular stereo datasets. Holopix50k is by-far the largest stereo dataset with top 3 scores in resolution, entropy and SR-metric}
\label{table:cross_dataset_metrics}
\begin{tabular}{lcccccc}
\hline
Dataset & Image Pairs & Resolution ($\uparrow$) & Entropy ($\uparrow$) & BRISQUE ($\downarrow$) & SR-metric ($\uparrow$) & ENIQA ($\downarrow$) \\
\hline
{\it KITTI 2012} \protect{\cite{kitti2012}} & 389 & 0.46 ($\pm 0.00$) Mpx & 7.12 ($\pm 0.30$) & {\bf 17.49} ($\pm 6.56$) & \underline{7.15} ($\pm 0.63$) & 0.097 ($\pm 0.028$) \\
{\it KITTI 2015} \protect{\cite{kitti2015}} & 400 & 0.47 ($\pm 0.00$) Mpx & 7.06 ($\pm 0.00$) & 23.79 ($\pm 5.81$) &  7.06 ($\pm 0.51$) & 0.169 ($\pm 0.030$) \\
{\it Middlebury} \protect{\cite{middlebury}} & 65 & {\bf 3.59} ($\pm 2.06$) Mpx & {\bf 7.55} ($\pm 0.20$) & 26.85 ($\pm 13.30$) &  6.01 ($\pm 1.08$) & 0.270 ($\pm 0.120$) \\
{\it ETH3D} \protect{\cite{eth3d}} & 47 & 0.38 ($\pm0.08$) Mpx & 7.24 ($\pm0.43$) & 27.95 ($\pm12.06$) & 5.99 ($\pm1.52$) & 0.195 ($\pm0.073$) \\
{\it Flickr1024} \protect{\cite{flickr1024}} & {1024} & {0.73} ($\pm0.33$) Mpx & 7.23 ($\pm0.64$) & 19.40 ($\pm13.77$) & 7.12 ($\pm0.67$) & {\bf 0.065} ($\pm0.073$) \\
{\it WSVD} \protect{\cite{wsvd}}\tnote{2} & \underline{10250} & 0.77 ($\pm0.29$) Mpx & 7.13 ($\pm0.59$) & 44.93 ($\pm12.66$) & 4.98 ($\pm1.54$) & 0.293 ($\pm0.112$) \\
{\it Holopix50k SD} & 13068 & 0.23 ($\pm0.00$) Mpx & \underline{7.42} ($\pm0.40$) & \underline{19.34} ($\pm11.94$) & {\bf 8.19} ($\pm0.95$) & \underline{0.069} ($\pm0.079$) \\
{\it Holopix50k HD} & 36300 & \underline{0.92} ($\pm0.00$) Mpx & 7.36 ($\pm0.50$) & 19.54 ($\pm11.77$) & 6.62\tnote{1} ($\pm0.98$) & 0.212 ($\pm0.116$) \\
\hline
{\it Holopix50k} & \textbf{49368} & 0.74 ($\pm0.30$) Mpx & 7.38 ($\pm0.48$) & 19.49 ($\pm11.81$) & 7.28\tnote{1} ($\pm1.24$) & 0.174 ($\pm0.125$) \\
\hline
\end{tabular}
\begin{tablenotes}
    \item Table adapted from \protect{\cite{flickr1024}}. \textbf{Bold} values indicate best scores, \underline{underlined} values indicate second-best scores.
    \item For all metrics, higher values are better, except BRISQUE and ENIQA.
    \item[1] Only {\it left} images from Holopix HD were used to compute SR-metric due to computation constraints. We assume trivial perceptual differences between {\it left} and {\it right} images.
    \item[2] Only the middle frame per clip is considered for our analysis, similar to the original paper \protect{\cite{wsvd}}.
\end{tablenotes}
\end{threeparttable}
}
\end{adjustbox}
\end{center}
\end{table}
\setlength{\tabcolsep}{1.8pt}

%% file: sections/5_experimentation/5_experimentation.tex
\section{Experiments with Holopix50k}

\input{sections/5_experimentation/5_1_stereo_sr_comparison}
\input{sections/5_experimentation/5_2_monodepth_comparison}
\input{sections/5_experimentation/5_3_our_internal_experiments}

%% file: sections/5_experimentation/5_1_stereo_sr_comparison.tex
\subsection{Stereo Super-Resolution}

In this section, we demonstrate how the Holopix50k dataset helps boost the performance on the current state-of-the-art method, namely PASSRNet \cite{passrnet2019longguang}, for stereo super-resolution. The PASSRNet model is originally trained on the Flickr1024 \cite{flickr1024} training dataset. We re-train it for the 4x super-resolution task in a variety of experiments using the training code provided by the authors. For these experiments, we use bicubic downsampling \cite{bicubic_downsampling} to create the low-resolution inputs needed to maintain parity with the Flickr1024 training. We compare the 4x super-resolved images to their original counterparts to calculate the PSNR and SSIM full-reference metrics for evaluating the performance of the models.

\input{tables/fig_qualitative_stereo_sr}

Figure \ref{fig:qualitative_stereo_sr} shows the qualitative results of the 4x super-resolution (SR) task on test images from various datasets. On visual inspection, it is evident that our Holopix50k model can resolve fine details and textures much better than both the Flickr1024 model and standard bicubic upsampling \cite{bicubic_interpolation}.

\input{tables/table_stereo_sr}

From Table \ref{table:stereo_sr_results}, we observe that using only 1k images from our dataset does not outperform the Flickr1024 model. However, when we start increasing the number of training samples, we note that our models outperform the Flickr1024 model on all test sets under consideration, except Flickr1024's own test set. These results show that our data improves generalization and prevents over-fitting. The model trained on the full Holopix50k training set outperforms all other models, further validating the need for such a large-scale stereo image dataset.

%% file: tables/fig_qualitative_stereo_sr.tex
\begin{figure}[!htb]
\centering
\begingroup
\setlength{\tabcolsep}{1pt}
    \scriptsize
    \begin{threeparttable}
    \begin{tabular}{ccccc}
    
    \subfloat{\includegraphics[width = 0.185\linewidth]{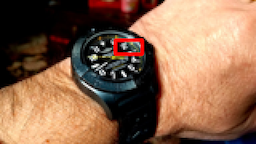}} &
    \subfloat{\includegraphics[width = 0.185\linewidth]{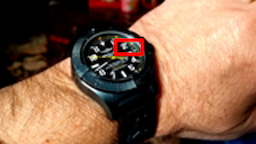}} &
    \subfloat{\includegraphics[width = 0.185\linewidth]{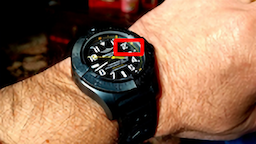}} &
    \subfloat{\includegraphics[width = 0.185\linewidth]{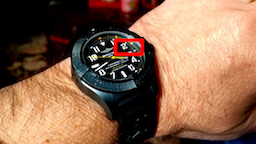}} &
    \subfloat{\includegraphics[width = 0.185\linewidth]{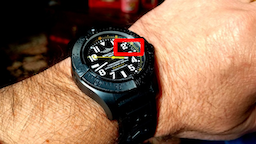}} \\ [-3.5ex]
    \subfloat{\includegraphics[width = 0.185\linewidth]{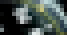}} &
     \subfloat{\includegraphics[width= 0.185\linewidth]{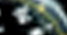}} &
    \subfloat{\includegraphics[width = 0.185\linewidth]{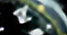}} &
    \subfloat{\includegraphics[width = 0.185\linewidth]{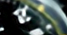}} &
    \subfloat{\includegraphics[width = 0.185\linewidth]{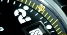}} \\ 
    
     Input &  Bicubic  & Flickr1024\tnote{*} & Holopix50k\tnote{*} & Ground Truth \\
     Holopix50k & 23.68/0.74 & 25.30/0.80 & 25.81/0.82 & \\
    
    \subfloat{\includegraphics[width = 0.185\linewidth]{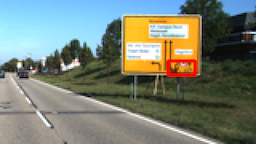}} &
    \subfloat{\includegraphics[width = 0.185\linewidth]{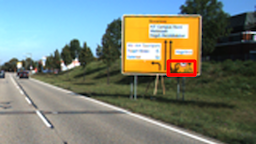}} &
    \subfloat{\includegraphics[width = 0.185\linewidth]{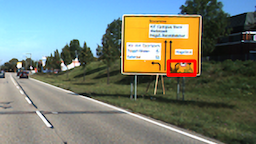}} &
    \subfloat{\includegraphics[width = 0.185\linewidth]{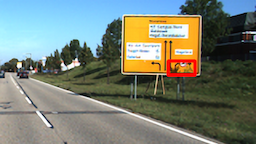}} &
    \subfloat{\includegraphics[width = 0.185\linewidth]{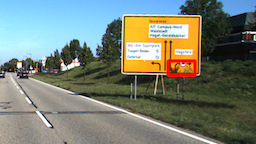}} \\ [-3.5ex]
    \subfloat{\includegraphics[width = 0.185\linewidth]{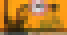}} &
     \subfloat{\includegraphics[width= 0.185\linewidth]{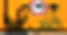}} &
    \subfloat{\includegraphics[width = 0.185\linewidth]{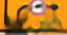}} &
    \subfloat{\includegraphics[width = 0.185\linewidth]{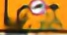}} &
    \subfloat{\includegraphics[width = 0.185\linewidth]{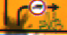}} \\ 
    
     Input &  Bicubic  & Flickr1024\tnote{*} & Holopix50k\tnote{*} & Ground Truth \\
     KITTI & 23.61/0.72 & 25.71/0.78 & 25.73/0.78 & \\
    \subfloat{\includegraphics[width = 0.185\linewidth]{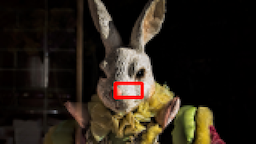}} &
    \subfloat{\includegraphics[width = 0.185\linewidth]{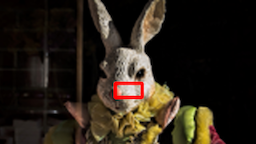}} &
    \subfloat{\includegraphics[width = 0.185\linewidth]{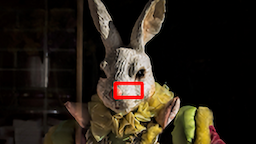}} &
    \subfloat{\includegraphics[width = 0.185\linewidth]{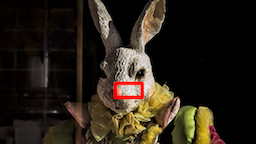}} &
    \subfloat{\includegraphics[width = 0.185\linewidth]{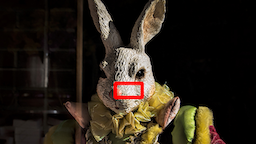}} \\ [-3.5ex]
    \subfloat{\includegraphics[width = 0.185\linewidth]{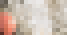}} &
     \subfloat{\includegraphics[width= 0.185\linewidth]{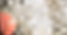}} &
    \subfloat{\includegraphics[width = 0.185\linewidth]{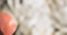}} &
    \subfloat{\includegraphics[width = 0.185\linewidth]{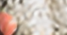}} &
    \subfloat{\includegraphics[width = 0.185\linewidth]{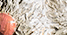}} \\
    Input &  Bicubic  & Flickr1024\tnote{*} & Holopix50k\tnote{*} & Ground Truth \\
    Flickr1024 & 24.84/0.85 & 25.55/0.88 & 25.89/0.89 & \\
    
    \subfloat{\includegraphics[width = 0.185\linewidth]{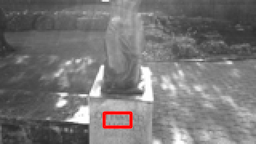}} &
    \subfloat{\includegraphics[width = 0.185\linewidth]{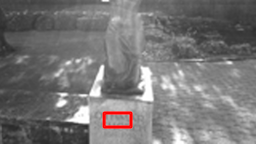}} &
    \subfloat{\includegraphics[width = 0.185\linewidth]{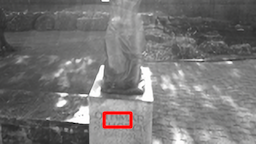}} &
    \subfloat{\includegraphics[width = 0.185\linewidth]{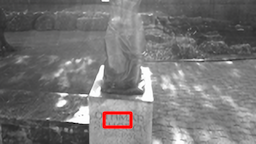}} &
    \subfloat{\includegraphics[width = 0.185\linewidth]{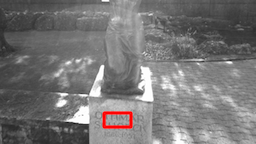}} \\ [-3.5ex]
    \subfloat{\includegraphics[width = 0.185\linewidth]{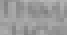}} &
     \subfloat{\includegraphics[width= 0.185\linewidth]{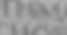}} &
    \subfloat{\includegraphics[width = 0.185\linewidth]{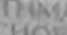}} &
    \subfloat{\includegraphics[width = 0.185\linewidth]{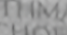}} &
    \subfloat{\includegraphics[width = 0.185\linewidth]{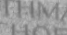}} \\
    Input &  Bicubic  & Flickr1024\tnote{*} & Holopix50k\tnote{*} & Ground Truth \\
    ETH3D & 28.96/0.83 & 30.48/0.87 & 30.93/0.88 & \\
    \end{tabular}
    \begin{tablenotes}
    \item[*] Refers to PASSRNet \protect{\cite{passrnet2019longguang}} trained on the respective datasets
    \end{tablenotes}
    \end{threeparttable}
    \endgroup
    \caption{Qualitative results of different super-resolution methods for 4x SR on images from a variety of datasets. Note that results produced by PASSRNet \protect{\cite{passrnet2019longguang}} trained on Holopix50k have the sharpest results and highest PSNR / SSIM scores}
    \label{fig:qualitative_stereo_sr}
\end{figure}

%% file: tables/table_stereo_sr.tex
\setlength{\tabcolsep}{4pt}
\begin{table}
\begin{center}
\begin{adjustbox}{width=1\textwidth}
{\renewcommand{\arraystretch}{1.4}
\begin{threeparttable}
\caption{PSNR (dB) and SSIM values from training PASSRNet \protect{\cite{passrnet2019longguang}} on various datasets for 4x SR with 80 training epochs}
\label{table:stereo_sr_results}
\begin{tabular}{lccccc}
\hline\noalign{\smallskip}
Dataset & KITTI2015 (Test) & Middlebury2014 (Test) & Flickr1024 (Test)& ETH3D (Test) & Holopix50k (Test)  \\
\noalign{\smallskip}
\hline
\noalign{\smallskip}
\textit{Flickr1024 (Train)} & 25.62 / 0.79 & 36.54 / 0.94 & {\bf23.25 / 0.72}  & 31.94 / 0.88 & 26.96 / 0.79 \\
\textit{Holopix50k (1K)} & 24.97 / 0.76 & 34.93 / 0.92 & 22.29 / 0.66 & 31.13 / 0.86 & 26.15 / 0.77 \\
\textit{Holopix50k (5K)} & 25.58 / 0.79 & 36.01 / 0.93 & 22.83 / 0.70 & 32.06 / 0.88 & 27.10 / 0.80 \\
\textit{Holopix50k (10K)} & \underline{25.71 / 0.79} & \underline{36.55 / 0.94} & 22.98 / 0.71  & \underline{32.29 / 0.88} & \underline{27.34 / 0.81} \\
\textit{Holopix50k (Train)} & {\bf25.78 / 0.79} & {\bf36.92 / 0.94} & \underline{23.13 / 0.71}  & {\bf32.64 / 0.89} & \textbf{27.56 / 0.81} \\
\hline
\end{tabular}
\begin{tablenotes}
    \item \textbf{Bold} values indicate best scores, \underline{underlined} values indicate second-best scores. For all metrics, higher values are better.
\end{tablenotes}
\end{threeparttable}
}
\end{adjustbox}
\end{center}
\end{table}
\setlength{\tabcolsep}{1.8pt}

%% file: sections/5_experimentation/5_2_monodepth_comparison.tex
\subsection{Self-supervised Monocular Depth Estimation}

We use the Holopix50k dataset to fine-tune a self-supervised monocular depth estimation model, namely Monodepth2 \cite{monodepth2}. The model is originally trained on the KITTI dataset \cite{kitti2012} and predicts a dense depth map from a single image. We use the training code provided by the authors with minor modifications for compatibility with our dataset. To compute the image reconstruction loss for training, we warp images using disparity-based backward mapping \cite{backward_map} as we do not have camera parameters or pose information.

\input{tables/fig_monodepth_qual}

Figure \ref{fig:monodepth_qualitative} shows qualitative results on images from the Middlebury and MPI Sintel datasets. The vanilla model produces blurry depth maps with only a few of the fine details from the original images intact. Results from fine-tuning on our dataset are sharper and maintain edges from the original images more consistently on both datasets.

\input{tables/table_monodepth_quant_new}

Table \ref{table:monodepth_quantitative_new} shows the quantitative results of running the original model and multiple fine-tuned models on the Middlebury and MPI Sintel test sets. We compare the disparity value by applying a per-image normalization and median scaling \cite{zhou2017unsupervised}. On the Middlebury dataset, our models perform better in all four error metrics. Specifically, the relative squared error is less than half of the original model, even with using 1k images from our dataset. On the MPI Sintel test set, the metrics are distributed between our models, with all the errors being significantly lower than the original KITTI model. From these results, we can conclude that our dataset helps a model trained on road scenes (from KITTI) generalize to entirely new scenarios that are also not present in our dataset.

%% file: tables/fig_monodepth_qual.tex
\begin{figure}[!htb]
\centering
\begingroup
\setlength{\tabcolsep}{1pt}
    \scriptsize
    \begin{tabular}{ccc@{\hskip0.05\linewidth}ccc}
    
    \subfloat{\includegraphics[width=0.15\linewidth, height = 1.25 cm]{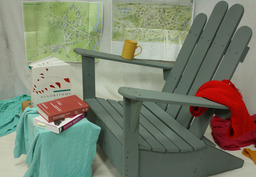}} &
    \subfloat{\includegraphics[width=0.15\linewidth, height = 1.25 cm]{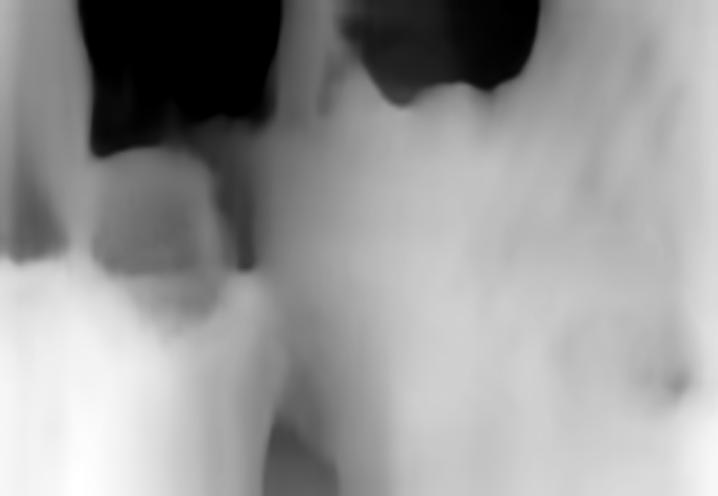}} &
    \subfloat{\includegraphics[width=0.15\linewidth, height = 1.25 cm]{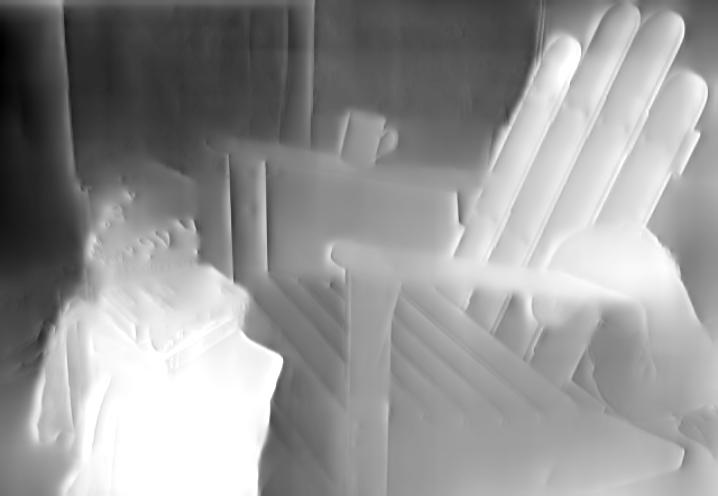}} &
    \subfloat{\includegraphics[width=0.15\linewidth, height = 1.25 cm]{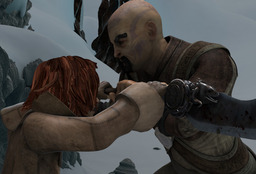}} &
    \subfloat{\includegraphics[width=0.15\linewidth, height = 1.25 cm]{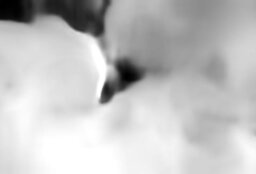}} &
    \subfloat{\includegraphics[width=0.15\linewidth, height = 1.25 cm]{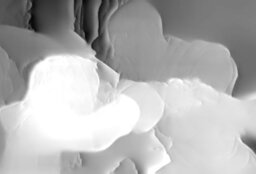}}
    \\
    [-3.5ex]
    \subfloat{\includegraphics[width=0.15\linewidth, height = 1.25 cm]{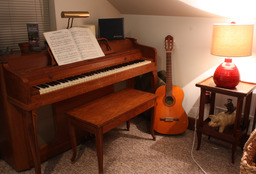}} &
    \subfloat{\includegraphics[width=0.15\linewidth, height = 1.25 cm]{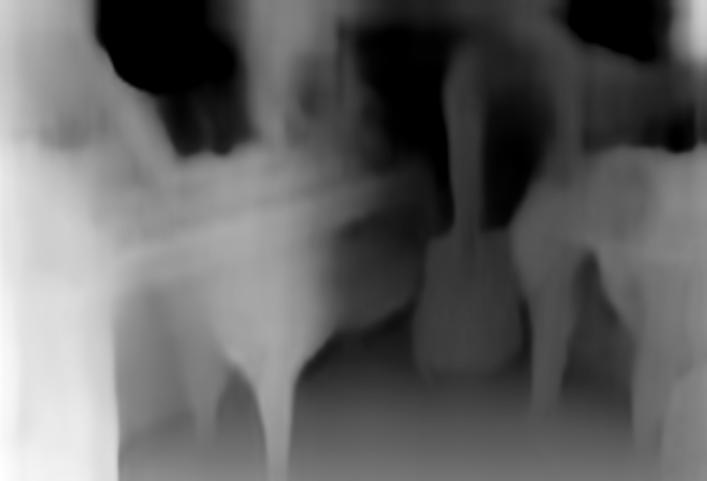}} &
    \subfloat{\includegraphics[width=0.15\linewidth, height = 1.25 cm]{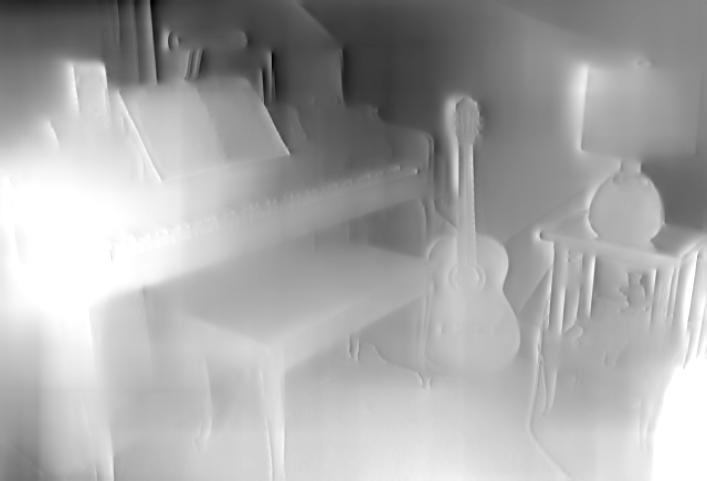}} &
    \subfloat{\includegraphics[width=0.15\linewidth, height = 1.25 cm]{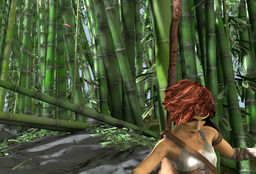}} &
    \subfloat{\includegraphics[width=0.15\linewidth, height = 1.25 cm]{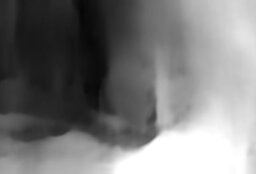}} &
    \subfloat{\includegraphics[width=0.15\linewidth, height = 1.25 cm]{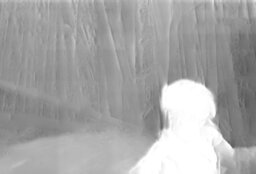}}    \\
    [-3.5ex]
    \subfloat{\includegraphics[width=0.15\linewidth, height = 1.25 cm]{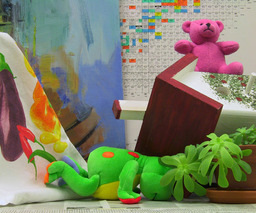}} &
    \subfloat{\includegraphics[width=0.15\linewidth, height = 1.25 cm]{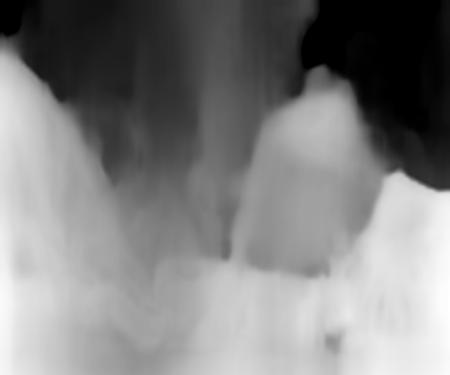}} &
    \subfloat{\includegraphics[width=0.15\linewidth, height = 1.25 cm]{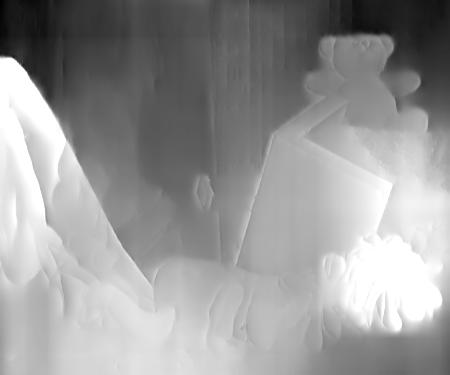}} &
    \subfloat{\includegraphics[width=0.15\linewidth, height = 1.25 cm]{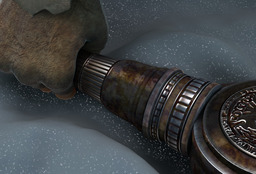}} &
    \subfloat{\includegraphics[width=0.15\linewidth, height = 1.25 cm]{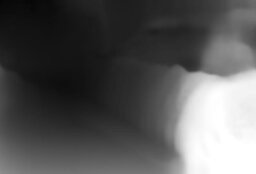}} &
    \subfloat{\includegraphics[width=0.15\linewidth, height = 1.25 cm]{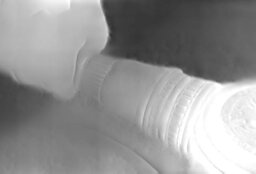}}    \\
        [-3.5ex]
    \subfloat{\includegraphics[width=0.15\linewidth, height = 1.25 cm]{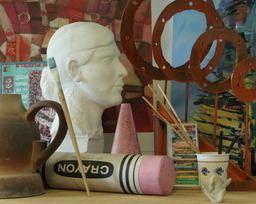}} &
    \subfloat{\includegraphics[width=0.15\linewidth, height = 1.25 cm]{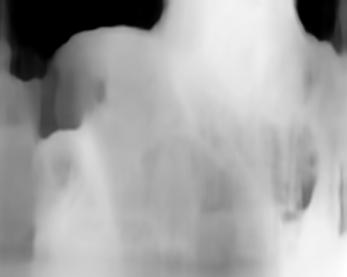}} &
    \subfloat{\includegraphics[width=0.15\linewidth, height = 1.25 cm]{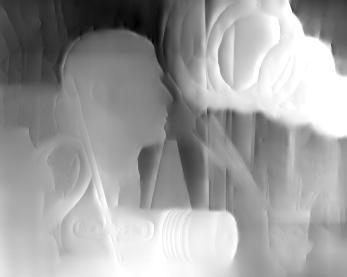}} &
    \subfloat{\includegraphics[width=0.15\linewidth, height = 1.25 cm]{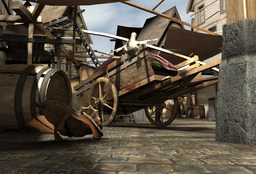}} &
    \subfloat{\includegraphics[width=0.15\linewidth, height = 1.25 cm]{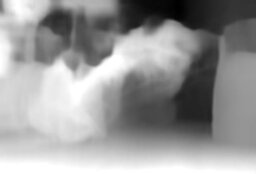}} &
    \subfloat{\includegraphics[width=0.15\linewidth, height = 1.25 cm]{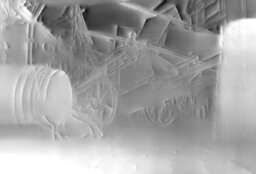}} 
    \\
    Middlebury & (a) & (b) & MPI Sintel & (a) & (b)
    \end{tabular}
    \endgroup
    \caption{Disparity maps produced by Monodepth2 \protect{\cite{monodepth2}} models on samples from the Middlebury \protect{\cite{middlebury}} (Left) and MPI Sintel \protect{\cite{butler2012mpi}} (Right) datasets, respectively. (a) Trained on KITTI, (b) Trained on Holopix50k}
    \label{fig:monodepth_qualitative}
\end{figure}

%% file: tables/table_monodepth_quant_new.tex
\setlength{\tabcolsep}{4pt}
\begin{table}
\begin{center}
\begin{adjustbox}{width=1\textwidth}
{\renewcommand{\arraystretch}{1.4}
\begin{threeparttable}
\caption{Disparity comparison of running Monodepth2 \protect{\cite{monodepth2}} models on the Middlebury visual benchmark and MPI Sintel test tests, respectively}
\label{table:monodepth_quantitative_new}
\begin{tabular}{lcccccccc}
\hline\noalign{\smallskip}
\multirow{2}{*}{Dataset} &
\multicolumn{4}{c}{Middlebury Test} &
\multicolumn{4}{c}{MPI Sintel Test} \\
& Abs Rel & Sq Rel & RMSE & RMSE log & 
Abs Rel & Sq Rel & RMSE & RMSE log \\
\noalign{\smallskip}
\hline
\noalign{\smallskip}
\textit{KITTI} & 0.590 & 15.968 & 20.891 & 0.308 & 0.337 & 31.726 & 41.597 & 0.173 \\
\textit{Holopix50k (1K) FT} & 0.421 & 7.528 & 14.224 & 0.203 & 0.239 & 17.062 & 25.882 & 0.114 \\
\textit{Holopix50k (5K) FT} & 0.420 & 7.387 & 14.136 & 0.202 & \textbf{0.235} & \textbf{16.474} & \textbf{25.300} & \underline{0.114} \\
\textit{Holopix50k (10K) FT} & \textbf{0.416} & \textbf{7.343} & \textbf{14.100} & \textbf{0.201} & 0.237 & 17.265 & 25.668 & 0.114 \\
\textit{Holopix50k (Train) FT} & \underline{0.417} & \underline{7.373} & \underline{14.112} & \underline{0.201} & \underline{0.235} & \underline{16.810} & \underline{25.302} & \textbf{0.113} \\
\hline
\end{tabular}

\begin{tablenotes}
    \item \textbf{Bold} values indicate best scores, \underline{underlined} values indicate second-best scores. For all metrics, lower values are better. \textit{FT} refers to fine-tuned model originally trained on the KITTI dataset.
\end{tablenotes}

\end{threeparttable}
}
\end{adjustbox}
\end{center}
\end{table}
\setlength{\tabcolsep}{1.8pt}

%% file: sections/5_experimentation/5_3_our_internal_experiments.tex
\label{sec:internal_training}
\subsection{Disparity Models Trained using Holopix50k}

In this section, we briefly describe models we have trained using the Holopix50k dataset for different disparity estimation tasks and share qualitative results. Note that we do not go into the implementation details of any of the networks or their training procedures. This section aims to motivate a variety of practical use cases and applications.

\subsubsection{Stereo Disparity Estimation.}

We use the Holopix50k dataset to train a stereo disparity estimation network for mobile inference. We jointly train a forward warping network as a part of our training pipeline. We use the disparity maps generated via semi-global block-matching \cite{sgbm} with additional filtering as supervised input during training. Our network follows a U-Net like architecture \cite{journals/corr/RonnebergerFB15} and has a computation footprint of {{$\scriptstyle\mathtt{\sim}$}340k} parameters ({$\scriptstyle\mathtt{\sim}$}1.5 GFLOPS). When using our Holopix50k dataset for this task, we observe that results have sharp edge detail (See Figure \ref{fig:inhouse_models}) and maintain stereo consistency. Our network also generalizes very well to a variety of real-world as well as synthetic scenarios. Images uploaded to Holopix are passed through this network to generate disparity maps. The final multi-view imagery users see on Holopix is synthesized using a combination of the original images and the generated disparity.

\input{tables/fig_our_disp_models}

\subsubsection{Real-time Disparity Estimation.}
We follow the same training pipeline and methodology as our stereo disparity estimation network to train a real-time disparity estimation network. The real-time network generates a single disparity map corresponding to both the left and right images. This network is optimized to have a computation footprint of {{$\scriptstyle\mathtt{\sim}$}15k} parameters ({$\scriptstyle\mathtt{\sim}$}0.15 GFLOPS). Though results are significantly smoother and have fewer edge details as seen in Figure \ref{fig:inhouse_models}, user testing demonstrates its feasibility for real-time mobile use-cases such as Lightfield camera previews, video calling, and video playback.

\subsubsection{Monocular Depth Estimation.}

We also train a monocular depth estimation network that is semi-supervised using a combination of the Holopix50k and Megadepth datasets \cite{DBLP:journals/corr/abs-1804-00607_megadepth}. The dense depth maps required in training are pseudo-labeled using our stereo neural network and the Megadepth model, respectively. We model this as a generative image translation task conditioned by stereo depth information using an architecture similar to Pix2pix \cite{DBLP:journals/corr/IsolaZZE16} with the PatchGAN discriminator \cite{DBLP:journals/corr/IsolaZZE16}. When using only the Megadepth dataset, we observe the model does not perform well on close-up to mid-range scenes, and scenes containing human-like subjects. By using our sufficiently large and diverse data set, our model learns to generate relative depths quite well from far away to close-up scenes, which has proven to be useful in a variety of practical scenarios. From Figure \ref{fig:inhouse_models}, we can see that the model correctly identifies the various depth layers present in the scene without any stereo supervision.

%% file: tables/fig_our_disp_models.tex
\begin{figure}[!htb]
\centering
\begingroup
\setlength{\tabcolsep}{1pt}
    \scriptsize
    \begin{tabular}{ccccc}
    
    \subfloat{\includegraphics[width=0.19\textwidth]{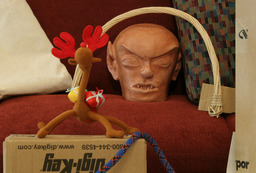}} &
    \subfloat{\includegraphics[width=0.19\textwidth]{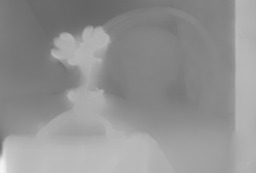}} &
    \subfloat{\includegraphics[width=0.19\textwidth]{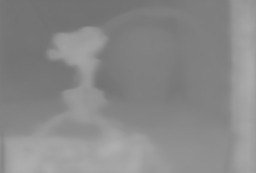}} &
    \subfloat{\includegraphics[width=0.19\textwidth]{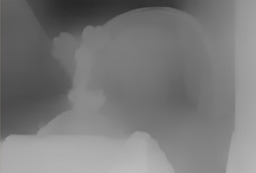}} & 
    \subfloat{\includegraphics[width=0.19\textwidth]{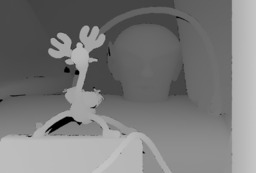}} 
    \\
    [-3.5ex]
    \subfloat{\includegraphics[width=0.19\textwidth]{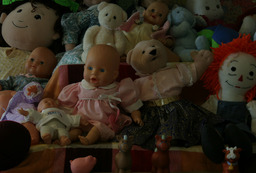}} &
    \subfloat{\includegraphics[width=0.19\textwidth]{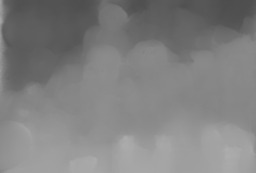}} &
    \subfloat{\includegraphics[width=0.19\textwidth]{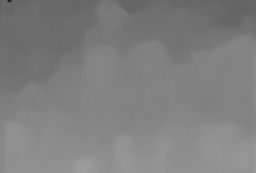}} &
    \subfloat{\includegraphics[width=0.19\textwidth]{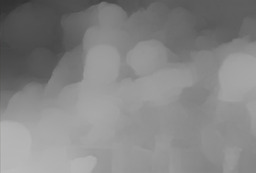}} & 
    \subfloat{\includegraphics[width=0.19\textwidth]{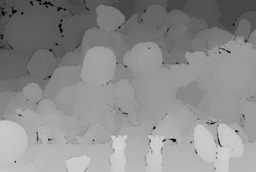}}
    \\
    Left Image & (a) & (b) & (c) & Ground Truth \\
    \end{tabular}
    \endgroup
    \caption{Left image disparity maps produced by our models on samples from the Middlebury dataset \protect{\cite{middlebury}}. (a) shows left disparity from our stereo disparity model, (b) shows result of our real-time disparity estimation, and (c) shows relative depth from our monocular depth estimation model}
    \label{fig:inhouse_models}
\end{figure}

%% file: sections/6_future_work.tex
\section{Future Work}

The Holopix50k dataset is our first release of a large-scale dataset scraped from the Holopix platform. As the platform continues to grow and increase its user-base, we plan to continue supporting the dataset with multiple future iterations. We also plan to share more details of the models trained using the Holopix50k dataset that we briefly cover in Section \ref{sec:internal_training}. Finally, since we have used the output disparity maps from our stereo disparity network as a reference for filtering the Holopix50k dataset, releasing these disparity maps as a pseudo-labeled ground truth dense disparity is another direction we can consider, based on the interest from the research community.

%% file: sections/7_conclusion.tex
\section{Conclusion}

In this work, we present \textit{Holopix50k}, a large-scale in-the-wild stereo image dataset that contains a variety of diverse scene types and semantic image content. To the best of our knowledge, this is the largest publicly released stereo image dataset collected from a mobile social media application. We showcase the superiority of our dataset, as a whole as well as its subsets, compared to other stereo datasets. We also demonstrate how using our dataset improves the performance of stereo image tasks, such as stereo super-resolution and self-supervised monocular depth estimation. We also briefly describe the models we trained using this dataset and the tasks they perform. We hope that this dataset encourages more work in the field of in-the-wild stereo imagery, both in areas we have discussed, as well as new research areas that are yet to be explored.